\documentclass[10pt,twocolumn,letterpaper]{article}

\usepackage{cvpr}              

\usepackage{graphicx}
\usepackage{amsmath}
\usepackage{amssymb}
\usepackage{booktabs}

\usepackage{times}
\usepackage{epsfig}
\usepackage{graphicx}
\usepackage{amsmath}
\usepackage{amssymb}
\usepackage{booktabs} 
\usepackage{subcaption}
\usepackage{arydshln}
\usepackage{wasysym}
\usepackage{dsfont}

\usepackage[accsupp]{axessibility}  

\usepackage[ruled,vlined, linesnumbered]{algorithm2e}

\DeclareMathOperator*{\argmin}{arg\,min}

\usepackage[pagebackref=true,breaklinks=true,colorlinks,bookmarks=false]{hyperref}

\usepackage[capitalize]{cleveref}
\crefname{section}{Sec.}{Secs.}
\Crefname{section}{Section}{Sections}
\Crefname{table}{Table}{Tables}
\crefname{table}{Tab.}{Tabs.}

\def\cvprPaperID{9791} 
\def\confName{CVPR}
\def\confYear{2022}

\begin{document}

\title{Unsupervised Learning of Debiased Representations with Pseudo-Attributes}

\author{
Seonguk Seo$^1$ \qquad Joon-Young Lee$^{3}$ \qquad Bohyung Han$^{1,2}$ \vspace{0.1cm} \\ 
 $^{1}$ECE \& $^{1}$ASRI \& $^{1,2}$IPAI, Seoul National University~~~~$^3$Adobe Research\\
 {\tt\small \{seonguk, bhhan\}@snu.ac.kr \quad jolee@adobe.com}
}
\maketitle

\begin{abstract}
Dataset bias is a critical challenge in machine learning since it often leads to a negative impact on a model due to the unintended decision rules captured by spurious correlations.
Although existing works often handle this issue based on human supervision, the availability of the proper annotations is impractical and even unrealistic.
To better tackle the limitation, we propose a simple but effective unsupervised debiasing technique.
Specifically, we first identify pseudo-attributes based on the results from clustering performed in the feature embedding space even without an explicit bias attribute supervision. 
Then, we employ a novel cluster-wise reweighting scheme to learn debiased representation; the proposed method prevents minority groups from being discounted for minimizing the overall loss, which is desirable for worst-case generalization.
The extensive experiments demonstrate the outstanding performance of our approach on multiple standard benchmarks, even achieving the competitive accuracy to the supervised counterpart.
The source code is available at our project page\footnote{\url{https://github.com/skynbe/pseudo-attributes}}.
\end{abstract}



\section{Introduction}

\begin{figure}
	\begin{center}
        \includegraphics[width=0.85\linewidth]{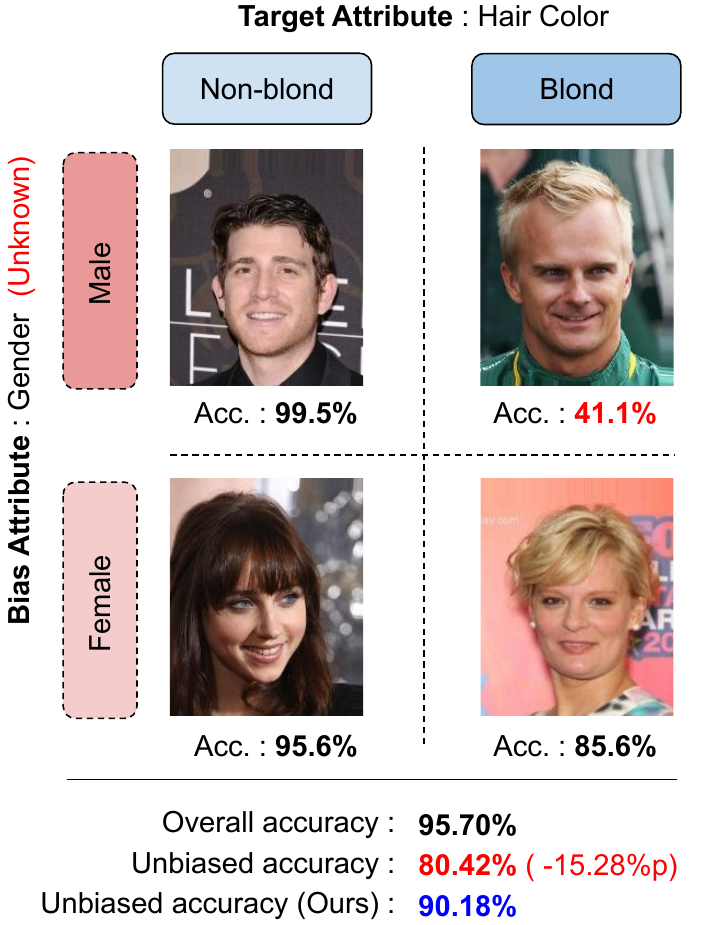}
        \vspace{-0.2cm}
	\end{center}
	\caption{Representative examples on the CelebA dataset for the problem that we focus on.
	Since most of the people with blond hair are women, \textit{hair color} attribute has a spurious correlation with \textit{gender} attribute. 
	Thus, when trained to classify the \textit{hair color}, a network captures the unintended decision rule using \textit{gender}, leading to poor worst-group and unbiased accuracies, despite its high overall accuracy.
	Our model aims to learn debiased representation, which gives better worst-group and unbiased accuracies, especially when the bias information is unavailable.
	}
	\label{fig:teasure}
	\vspace{-0.1cm}    
\end{figure}

Deep neural networks have achieved impressive performance by minimizing the average loss on training datasets.
Although we typically adopt the empirical risk minimization framework as a training objective, it is sometimes problematic due to dataset bias leading to significant degradation of worse-case generalization performance as discussed in \cite{ben2013robust, zemel2013learning, li2017deeper, geirhos2020shortcut, zhang2020adaptive}.
This is because models do not always learn what we expect, but, to the contrary, rather capture unintended decision rules from spurious correlations.
For example, on the Colored MNIST dataset~\cite{li2019repair, LfF, Rebias}, where each digit is highly correlated to a certain color, a network often learns the color patterns in images, not the digit information, while ignoring few conflicting samples.
Such an unintended rule works well on most of the training samples but incurs unexpected worst-case errors for the examples in minority groups, which makes the model unable to generalize on test environments with distribution shifts or robustness constraints.
Figure~\ref{fig:teasure} illustrates the problem that we mainly deal with in this paper.

To mitigate the bias issue, learning debiased representations from a biased dataset has received growing attention from machine learning community~\cite{li2019repair, wang2018learning, GroupDRO, seo2022information, hendricks2018women, Rebias, RUBi}.
However, most previous works rely on the explicit supervision or prior knowledge under the assumption of the presence of dataset bias.
This setting is problematic because it is challenging to identify what kinds of bias exist and which attributes involve spurious correlations without thorough analysis of model and dataset.
Note that, even with the information about dataset bias, the relevant annotations over all training examples is challenging.
Contrary to the supervised approaches, \cite{LfF, sohoni2020no} tackle a more challenging setting, where the bias information is unavailable, via failure-based learning or subgroup-based penalizing.

This paper presents a simple but effective unsupervised debiasing technique via feature clustering and cluster re-weighting.
We first observe that the examples with the same label for a certain attribute other than the target attribute tend to have similar representations in the feature space by the model trained sufficiently.
Based on this observation, we estimate bias \textit{pseudo-attributes} in an unsupervised manner from the clustering results within each class.
To exploit the bias pseudo-attributes for learning debiased representations, we introduce a reweighting scheme for the corresponding clusters, where each cluster has an importance weight depending on its size and task-specific accuracy.
This strategy encourages the minority clusters to participate in the optimization process actively, which is critical to improving worst-group generalization.
Despite its simplicity, our method turns out to be effective for debiasing without any supervision of bias information; it is even comparable to the supervised debiasing method. 
%
The main contributions of our work are summarized below:
 \begin{itemize} 
	\item[$\bullet$] We propose a simple but effective unsupervised debiasing approach, which requires no explicit supervision about spurious correlations across attributes.
    \item[$\bullet$] We introduce a technique to learn debiased representations by identifying bias pseudo-attributes via clustering and reweighting the corresponding clusters based on both their size and target loss.
    \item[$\bullet$] We provide extensive experimental results and achieve outstanding performance in terms of unbiased and worst-group accuracies, which are even as competitive as supervised debiasing methods.
\end{itemize}

The rest of this paper is organized as follows.
We review the prior research in Section~\ref{sec:related}.
Section~\ref{sec:method} presents the proposed framework for learning debiased representations, and Section~\ref{sec:experiments} demonstrates its effectiveness with extensive empirical analysis.
We conclude our paper in Section~\ref{sec:conclusion}.


\section{Related Work}
\label{sec:related}

\subsection {Bias in computer vision tasks}
\label{sub:bias}
Real-world datasets are inevitably biased due to their insufficiently controlled collection process, and consequently, deep neural networks often capture unintended correlations between the true labels and the spuriously correlated ones.
Measuring and mitigating the potential risks posed by dataset or algorithmic bias has been extensively investigated in various computer vision tasks~\cite{chu2021learning, torralba2011unbiased, li2019repair, johnson2017clevr, revisetool, RUBi, wang2019iccv}.
For example, VQA models frequently exploits statistical regularities between answer occurrences and patterns of questions while ignoring visual information~\cite{RUBi, clark2019don}.
Semantic segmentation models typically take advantage of scene context for pixel-level predictions of semantic labels~\cite{chu2021learning}.
To prevent using undesirable correlation in biased datasets, existing approaches often rely on human supervision for bias annotations and present several technical directions such as data augmentation~\cite{geirhos2018imagenet, zhang2020towards}, model ensembles~\cite{RUBi, clark2019don}, and {statistical regularizations~\cite{Rebias}.}
Such supervised debiasing techniques have been applied to various computer vision tasks by exploiting known application-specific bias information, including uni-modality of the dataset in visual question answering~\cite{RUBi}, stereotype textures in image recognition~\cite{geirhos2018imagenet}, temporal invariance in action recognition~\cite{li2018resound}, and demographic information in facial image recognition~\cite{GroupDRO, zhang2020towards, wang2019racial}.

\subsection {Handling distribution shifts}
Distribution shift has recently emerged as a critical challenge in machine learning, where the goal of the optimization is to learn a robust model in a test environment with a different distribution.
Distributionally robust optimization (DRO)~\cite{ben2013robust, gao2017wasserstein, duchi2016statistics, DRBO} has been proposed to improve the worst-case generalization performance over a family of target distributions, and has provided theoretical background of Group DRO~\cite{GroupDRO} and its variation~\cite{sohoni2020no}.
However, the objective of DRO often leads to an overly conservative model and results in performance degeneration on unseen environments~\cite{duchi2020distributionally, hu2018does}.
To relax the constraints for the uncertainty set of test distributions, some approaches pose additional assumptions.
For instance, the dataset consists of multiple groups with the shared properties and the uncertainty set is represented by a mixture of these groups.
This assumption is also used in robust federated learning~\cite{mohri2019agnostic, li2019fair}, algorithmic fairness~\cite{zemel2013learning, chouldechova2018frontiers, donini2018empirical}, and domain generalization~\cite{li2017deeper, carlucci2019domain, seo2019learning}.
Our framework also takes advantage of this assumption but does not rely on the supervision of group information.

\subsection{Debiasing via loss-based reweighting}
There exist several generic debiasing techniques via sample reweighting based on observed task-specific losses under the supervised environment~\cite{GroupDRO} or the unsupervised setting~\cite{LfF, sohoni2020no}.
Group DRO~\cite{GroupDRO} exploits the group information specified by the bias attributes and aims to improve the worst-group generalization performance.
On the other hand, Nam~\etal~\cite{LfF} employ the difference between the generalized and standard cross-entropy loss to capture the bias-alignment for sample reweighting while Sohoni~\etal~\cite{sohoni2020no} estimate subclass labels via clustering and utilize the information for distributionally robust optimization to mitigate hidden stratification.
Although the unsupervised approaches work well in small and artificial datasets such as MNIST, their performance improvement becomes marginal in real-world datasets including CelebA.
Our framework also belongs to unsupervised methods that do not rely on the bias information to learn debiased representations.


\section{Method}
\label{sec:method}

This section presents our debiasing technique via bias pseudo-attribute estimation and sample reweighting.

\subsection{Preliminaries}
\label{sub:preliminaries}
Let an example $\mathbf{x}$ be associated with a set of $m$ attributes $ \mathcal{A}:= \{a_1, ..., a_m\}$.
The goal of our model is to predict a target attribute $a_t \in \mathcal{A}$ by estimating the intended causation $p(a_t|\mathbf{x})$, which does not involve any undesirable correlation to other latent attributes, \ie, $p(a_t|\mathbf{x}) = p(a_t|\mathbf{x}, a_i)$, $\forall a_i  \in \mathcal{A} - \{a_t\}$.
On the other hand, spurious correlation indicates strong coincidence between two attributes $a_i, a_j \in  \mathcal{A}$; the conditional entropy $H(a_i|a_j)$ is close to zero and there exists  no causal relationship between them.
A machine learning algorithm is considered biased if a certain attribute $a_b \in \mathcal{A}$ has a spurious correlation with the target attribute $a_t$ and affects the prediction, \ie, $p(a_t|\mathbf{x}) \neq p(a_t|\mathbf{x}, a_b)$. 
Our approach performs debiasing by estimating groups in the dataset without supervision, where the group is defined by a pair of target and bias attributes, \eg, $g = (a_t, a_b)$.

\subsection{Observation}
\label{sub:observation}

If a bias attribute is highly correlated to a target attribute while being easy to learn, the model may ignore few conflicting examples and learn its decision rule based on the bias attributes with spurious correlations to maximize accuracy~\cite{LfF, sagawa2020investigation}.
To prevent this undesirable situation, a simple group upweighting or resampling strategies~\cite{sagawa2020investigation} are known to be effective while they work poorly in realistic scenarios, where the bias information is unknown during training.

To overcome this challenge, from our intuition, we analyze the feature semantics over the target and bias attributes.
We first na\"ively train a base model on the CelebA dataset to classify \textit{hair color}, and visualize the representation of the examples after convergence with a sufficient number of epochs ($T = 100$).
We select \textit{gender} as a bias attribute, but do not utilize any information of the bias attribute during training.
It turns out that, even without using the bias information during training, the examples drawn from certain groups, which are given by a combination of hair color and gender attribute values in this case, \eg, (male, non-blonde) and (female, blonde), are located closely in the feature space.
This observation implies that it is possible to identify bias {pseudo-attributes} by taking advantage of the embedding results even without attribute supervisions.
Our unsupervised debiasing framework is based on the capability to identify the bias pseudo-attributes via clustering.

\subsection{Formulation}

Suppose that training examples, $(\mathbf{x}, y)$, are drawn from a certain distribution $\tilde{P}$.
Given a loss function $\ell(\cdot)$ and model parameters $\theta$, the objective of the empirical risk minimization (ERM) is to optimize the following expected loss:
\begin{align}
  \min_{\theta} \ \mathbb{E}_{(\mathbf{x}, y) \sim {P}} \big[\ell( (\mathbf{x}, y); \theta) \big],
\end{align}
where ${P}$ is the empirical distribution over training data that approximates the true distribution $\tilde{P}$.
Although ERM generally works well, it tends to ignore examples in minority groups that conflict with bias attributes and implicitly assumes the consistency of the underlying distributions for training and testing data.
Consequently, the approach often leads to high unbiased and worst-group test error~\cite{duchi2016statistics, GroupDRO}.

Several distributionally robust optimization (DRO) techniques~\cite{ben2013robust, duchi2016statistics, DRBO} can be employed to tackle the dataset bias and distribution shift problems and maximize unbiased generalization accuracy.
They consider a particular uncertainty set $\mathcal{Q}_P$, which is close to the training distribution $P$, \eg, $\mathcal{Q}_P = \{Q : D_f[Q||P] \leq \delta \}$, where $D_f[\cdot||\cdot]$ indicates an $f$-divergence function\footnote{Let $P$ and $Q$ be probability distributions over a space $\Omega$, then \textit{f}-divergence is $D_f(P||Q) = \int_\Omega f \big(\frac{dP}{dQ} \big) dQ$.}.
To minimize the worst-case loss over the uncertainty distribution set $\mathcal{Q}_P$, DRO optimizes
\begin{align}
  \underset{\theta}{\vphantom{\sup}\min} \Bigl\{ \mathbb{R}_{ {\mathcal{Q}_P}}(\theta) := \sup_{Q \in {\mathcal{Q}_P}} \mathbb{E}_{(\mathbf{x}, y) \sim Q} \big[ \ell( (\mathbf{x}, y)); \theta \big] \Bigr\}.
\end{align}
However, this objective is overly pessimistic and makes the model consider the implausible worst cases~\cite{duchi2020distributionally, hu2018does}.

The group distributionally robust optimization, referred to as group DRO~\cite{GroupDRO} creates more realistic sets of possible test distributions by leveraging the prior knowledge of group information.
They assumes the training distribution $P$ is a mixture of $G$ groups, $P_g$, which is given by 
\begin{equation}
P_\mathcal{G} = \sum_{g \in \mathcal{G}} c_g{P_g}, ~~~c \in \Delta_G
\end{equation}
where $\mathcal{G} = \{1, ..., G\}$ and  $\Delta_G$ is a $(G-1)$-dimensional simplex.
Then the uncertainty set $\mathcal{Q}_P$ is defined by a set of all possible mixtures of these groups, \ie, $\mathcal{Q}_{P_\mathcal{G}} = \{\sum_{g \in \mathcal{G}} c_g P_g : c \in \Delta_G \}$.
Because $\mathcal{Q}_{P_\mathcal{G}}$ is a simplex, its optimum is achieved at a vertex, thus minimizing the worst-case risk of $\mathcal{Q}_{P_\mathcal{G}}$ is equivalent to
\begin{align}\label{eq:gdro}
   \underset{\theta}{\vphantom{\sup}\min} \Bigl\{{\mathbb{R}_\mathcal{G}}(\theta) :=  \underset{g \in \mathcal{G}}{\vphantom{\argmin}\max} \ \mathbb{E}_{(\mathbf{x}, y) \sim P_g} \big[ \ell((\mathbf{x}, y); \theta) \big] \Bigr\}.
\end{align}

Different from the group DRO setting, we do not know the group assignment for each training example.
Instead, we use the bias pseudo-attribute information, obtained by any clustering algorithm in the feature embedding space, to define groups.
Note that the clustering is performed with the representations given by the base model trained without debiasing, which is parameterized by $\Tilde{\theta}$.
Our goal is to alleviate dataset bias and maximize unbiased accuracy, and we need to consider all groups fairly for optimization.
To this end, we assign a proper importance weight, $\omega_k$, to the $k^\text{th}$ cluster,  where $k \in \mathcal{K} = \{1, ..., K\}$,
and the final objective of our framework is given by minimizing a weighted empirical risk as follows:%
\begin{align}
   \underset{\theta}{\vphantom{\sup}\min} \Bigl\{{\mathbb{R}_\mathcal{K}}(\theta) := \mathbb{E}_{(\mathbf{x}, y) \sim P} \Big[  \omega_{h((\mathbf{x},y);\Tilde{\theta})}   \ell((\mathbf{x}, y); \theta) \Big] \Bigr\},
   \label{eq:pgdro}
\end{align}
where $h((\mathbf{x},y);\Tilde{\theta})$ denotes the cluster membership of an example $(\mathbf{x}, y)$.
The details of the weight assignment method will be discussed next.

\begin{algorithm}[t]
\SetAlgoLined
\textbf{Require:} {step size $\eta_\theta$}, momentum $m$, training steps $T$, batch size $B$, the number of clusters $K$ \\
\vspace{0.1cm}
\textbf{Base model:}  \\
\textbf{Initialize} $\Tilde{\theta}$  \\
\For{$t=1,...,T$}{
Sample $(\mathbf{x}_i, y_i) \sim P$ for $i = 1, ..., B$\;
$\Tilde{\theta} \gets \Tilde{\theta} - \eta_\theta \sum_{i=1}^B \nabla \ell( (\mathbf{x}_i, y_i); {\Tilde{\theta}})$\;
}
\vspace{0.2cm}
\For{$k=1,...,K$}{
$P_k = \{(\mathbf{x}_n, y_n)~|~h((\mathbf{x}_n, y_n); \Tilde{\theta}) = k~~\text{for all}~n  \}$\;
$N_k = |P_k| $\;
}
\vspace{0.4cm}
\textbf{Target model:} \\
\textbf{Initialize} $\theta$ and $\omega_k$ for $k = 1, ..., K$\\
 \For{$t=1,...,T$}{
$\omega_k \gets (1-m)\omega_k + \frac{m}{N_k} {\mathbb{E}_{(\mathbf{x}, y) \sim {P_k}}[\ell( (\mathbf{x}, y); {\theta})]}$ \\~~~~~~~~~~~~~~~~~~~~~~~~~~~~~~~~~~~~~~~~~~~~~~~~~~for $k = 1, ..., K$\;

Sample $(\mathbf{x}_i, y_i) \sim P$ for $i = 1, ..., B$\;
$\alpha_i = \omega_{h((\mathbf{x}_i, y_i); \Tilde{\theta})}$\;
$\overline{\alpha}_{i} = {\alpha_{i}}  / {\sum_{i=1}^B \alpha_{i}} $\;
$\theta \gets \theta - \eta_\theta \sum_{i=1}^B \overline{\alpha}_{i} \nabla \ell( (\mathbf{x}_i, y_i); {\theta})$;
 }
 \caption{Debiasing with bias pseudo-attribute}
 \label{alg:algorithm}
\end{algorithm}
\vspace{0.3cm}

\subsection{Sample weighting with bias pseudo-attributes}
\label{sub:sample}
Based on our observation described in Section~\ref{sub:observation}, we first cluster training examples in each class on the feature embedding space defined by the base model optimized sufficiently, \eg, for 100 epochs using the standard classification loss.
We suppose that each cluster corresponds to a bias pseudo-attribute.
Among all clusters, we focus on the examples in the minority clusters, especially when they have high average losses.
A common failure case in the presence of dataset bias is incurred by ignoring specific subpopulation groups for minimizing the overall training loss, and minority clusters are prone to be ignored due to their sizes.
The problematic cases among the clusters are the ones that contain many bias-conflicting examples, having high losses, and thus result in poor worst-case errors.
If the minority clusters consist of mostly bias-aligned samples, they will apparently achieve high classification accuracy.

Therefore, to handle dataset bias issue, we should consider both scale and average difficulty (loss) of each cluster, unlike group DRO~\cite{GroupDRO} and George~\cite{sohoni2020no}, which focus only on the average loss.
We calculate the importance weight of each cluster by our reweighting scheme to train the target model, which is given by
\begin{align}
   \omega_k 
   &=  \frac {\mathbb{E}_{(\mathbf{x}, y) \sim {P_k}}[\ell( (\mathbf{x}, y); {\theta})]} {{N_k}^{}}
   \label{eq:importance} \nonumber \\
   &= \frac {\mathbb{E}_{(\mathbf{x}, y) \sim {P}} \big[ \ell( (\mathbf{x}, y); {\theta})~|~{h((\mathbf{x}, y);\Tilde{\theta})=k}  \big]^{}} { \sum_i \mathds{1}(h((\mathbf{x}_i, y_i); \Tilde{\theta}) = k)},
\end{align}
where $\theta$ and $\Tilde{\theta}$ indicates the parameters of the final and base models, respectively, $h(\cdot, \cdot)$ is a cluster membership function, and $\mathds{1}(\cdot)$ is an indicator function. 
Note that ${P_k}$ denotes the sample distribution of the $k^\text{th}$ cluster and $N_k$ is the number of samples in the $k^\text{th}$ cluster, where $k \in \mathcal{K} = \{1, ..., K\}$.

\subsection{Algorithm procedure}

Algorithm~\ref{alg:algorithm} presents the optimization procedure of the proposed framework.
We first n\"aively train a baseline network (line 4-7), parameterized by $\Tilde{\theta}$.
Then, we cluster all training examples based on the features extracted from the network to obtain the membership distribution $P_k$ and the size of cluster $N_k$ (line 8-11).
Based on the cluster assignments, we calculate the importance weight of each cluster $\omega_k$ using the target model, parameterized by $\theta$, where the weight is updated by exponential moving average at each iteration (line 15).
We finally use the normalized importance weight of each sample $\overline{\alpha}_i$ over a mini-batch to train the target model (line 18-19).


\begin{table*}[t]
\begin{center}
\caption{Unbiased and worst-group results in the existence of spurious correlation between target and bias attributes on the test split of the CelebA dataset.
LfF$^*$ denotes a variant of LfF~\cite{LfF}, which fine-tuned only the classification layer of a trained baseline model, for additional comparison to ours.
Bold and underline fonts indicate the first and second place among the compared approaches, respectively.
All experimental results are the average of thee runs.
}
\label{tab:celeba}
 \scalebox{0.85}{
\setlength\tabcolsep{6pt} \hspace{-0.25cm}
\begin{tabular}{c|c|ccc|c|c||ccc|c|c}
\toprule
\multicolumn{2}{c|}{}&\multicolumn{5}{c||}{Unbiased accuracy (\%)}  &\multicolumn{5}{c}{Worst-group accuracy (\%)} \vspace{0.1cm} \\
\cdashline{3-12} 
&&\multicolumn{4}{c|}{Unsupervised} & Supervised & \multicolumn{4}{c|}{Unsupervised} & Supervised \vspace{0.02cm} \\
    Target & Bias & Base & LfF & LfF$^*$ & {BPA (ours)} & Group DRO & Base & LfF & LfF$^*$ & {BPA (ours)} & Group DRO\\
\hline
Blond Hair& Gender &80.42&59.46&84.89&{\underline{90.18}}&\bf{91.39}		&41.02&34.23&57.96&{\underline{82.54}}&\bf{87.86}\\
Heavy Makeup& Gender &71.19&56.34&71.85&\bf{73.78}&{\underline{72.70}}		&17.35&{\underline{30.81}}&23.87&\bf{39.84}&21.36\\
Pale Skin& Gender &71.50&78.69&75.23&{\underline{90.06}}&\bf{90.55}		&36.64&57.38&43.26&\bf{88.60}&{\underline{87.68}}\\
Wearing Lipstick& Gender &73.90&53.79&73.84&\bf{78.28}&{\underline{78.26}}		&31.38&25.52&31.92&\bf{46.52}&{\underline{46.08}}\\
Young& Gender &78.19&45.99&79.58&{\underline{82.27}}&\bf{82.40}		&52.79&0.34&57.79&{\underline{74.33}}&\bf{76.29}\\
Double Chin& Gender &64.61&65.46&68.47&{\underline{82.92}}&\bf{83.19}	&21.33&28.19&28.24&{\underline{67.78}}&\bf{72.94}\\
Chubby& Gender &67.42&60.03&71.56&\bf{83.88}&{\underline{81.90}}				&24.30&7.60&34.09&{\underline{72.32}}&\bf{72.64}\\
Wearing Hat& Gender &93.53&84.56&94.81&{\underline{96.80}}&\bf{96.84}			&85.12&69.06&88.31&\bf{94.94}&{\underline{94.67}}\\
Oval Face& Gender &62.70&57.64&62.30&\bf{67.18}&{\underline{65.40}}		&29.15&7.40&36.00&{\underline{55.78}}&\bf{56.84}\\
Pointy Nose& Gender &62.10&42.20&63.83&{\underline{68.90}}&\bf{70.71}			&25.80&1.05&38.04&{\underline{52.48}}&\bf{63.76}\\
Straight Hair& Gender &70.28&39.57&72.84&\bf{79.18}&{\underline{77.04}}			&47.82&1.95&58.53&\bf{72.09}&{\underline{66.10}}\\
Blurry& Gender &73.05&76.70&77.52&\bf{88.93}&{\underline{87.05}	}			&45.68&43.81&52.35&\bf{84.10}&{\underline{82.06}}\\
Narrow Eyes& Gender &63.18&68.53&67.77&{\underline{76.39}}&\bf{76.72}			&27.01&31.81&38.53&\bf{73.24}&{\underline{71.47}}\\
Arched Eyebrows& Gender &69.72&56.17&71.87&{\underline{74.77}}&\bf{78.30}		&34.76&26.21&44.97&{\underline{54.36}}&\bf{69.44}\\
Bags Under Eyes& Gender &69.47&44.61&71.86&\bf{77.84}&{\underline{75.88}}		&41.65&0.06&49.10&{\underline{62.55}}&\bf{63.34}\\
Bangs& Gender &89.04&41.41&89.04&{\underline{93.94}}&\bf{94.45}				&76.91&3.18&82.37&\bf{92.21}&{\underline{92.12}}\\
Big Lips& Gender &60.87&46.74&62.15&\bf{66.50}&{\underline{63.70}}			&30.85&31.44&38.54&\bf{56.99}&{\underline{47.55}}\\
No Beard& Gender &73.11&60.12&73.13&\bf{79.58}&{\underline{77.86}}			&13.30&11.92&20.00&\bf{40.00}&{\underline{36.70}}\\
Receding Hairline& Gender &69.72&70.57&74.58&{\underline{84.95}}&\bf{85.15}		&35.69&32.10&45.53&{\underline{79.11}}&\bf{79.12}\\
Wavy Hair& Gender &73.10&48.00&74.53&\bf{79.89}&{\underline{79.65}}			&38.01&0.06&45.24&{\underline{65.74}}&\bf{66.79}\\
Wearing Earrings& Gender &72.17&59.35&74.17&\bf{84.57}&{\underline{83.50}}		&26.26&0.10&32.95&{\underline{72.81}}&\bf{75.24}\\
Wearing Necklace& Gender &55.04&58.64&57.21&\bf{68.96}&{\underline{62.89}}		&2.72&0.22&6.67&\bf{41.93}&{\underline{24.34}}\\
\hline
    \bf{Average} & Gender & 72.67 & 58.65 & 74.87 & \bf{81.74} & {\underline{80.87}} & 39.91 & 21.91&  47.88 &  \bf{69.84}  & {\underline{69.68}}  \\
\bottomrule
\end{tabular}
 }
\end{center}
\vspace{-0.2cm}
\end{table*}

\section{Experiments}
\label{sec:experiments}

\begin{table*}[t]
\begin{center}
\caption{Unbiased and worst-group results on the Waterbirds dataset.}
\vspace{-0.1cm}
\label{tab:waterbirds}
 \scalebox{0.85}{
\setlength\tabcolsep{6pt} \hspace{-0.2cm}
\begin{tabular}{c|c|ccc|c|c||ccc|c|c}
\toprule
\multicolumn{2}{c|}{}&\multicolumn{5}{c||}{Unbiased accuracy (\%)}  &\multicolumn{5}{c}{Worst-group accuracy (\%)} \vspace{0.1cm} \\
\cdashline{3-12}
&&\multicolumn{4}{c|}{Unsupervised} & Supervised & \multicolumn{4}{c|}{Unsupervised} & Supervised \\
    Target & Bias & Base & LfF & LfF$^*$ & BPA (ours) & Group DRO & Base & LfF & LfF$^*$ & BPA (ours) & Group DRO\\
\hline
Object & Place &84.63 & {85.48} & 84.57 & {\underline{87.05}}& {\bf{88.99}} & 62.39 &68.02&  61.68 & {\underline{71.39}} & {\bf{80.82}}\\
 Place & Object &87.99& {85.77} & 85.05 & {\underline{88.44}} & {\bf{89.20}} & 73.34 &62.37 &  60.00 & {\underline{79.16}} & {\bf{85.27}} \\
\bottomrule
\end{tabular}
 }
\end{center}
\vspace{-0.1cm}
\end{table*}

\begin{table}[t]
\begin{center}
\caption{Unbiased accuracy (\%) on the valid split of the Colored-MNIST dataset.}
\vspace{-0.1cm}
\label{tab:mnist}
\scalebox{0.85}{
\setlength\tabcolsep{6pt}
\hspace{-0.2cm}
\begin{tabular}{c|c|cc|c||c}
\toprule
&& \multicolumn{3}{|c||}{Unsupervised} & Supervised \\
     Target & Bias & Baseline & LfF & BPA (ours) & Group DRO\\
\hline
    Digit & Color & 74.48  & {85.15} & {\underline{85.26}} & {\bf{85.88}}  \\
    Color & Digit & {\bf{99.95}} & {\underline{99.91}} & 99.82 & 98.96 \\
\bottomrule
\end{tabular}
}
\end{center}
\vspace{-0.3cm}
\end{table}

\subsection{Dataset}
CelebA~\cite{CelebA} is a large-scale face dataset for face image recognition, containing 40 attributes for each image.
Following the previous works~\cite{GroupDRO, LfF}, we set \textit{hair color} and \textit{heavy makeup} as the target attribute.
Note that \textit{gender} attribute is spuriously correlated to the two attributes and employed as the bias attribute for worst-group accuracy evaluation in our experiment.
For more comprehensive results, we also consider the other 32 attributes as the target attributes.

Waterbirds~\cite{GroupDRO} is a synthesized dataset with 4,795 training examples, created by combining birds photographs from the Caltech-UCSD Birds-200-2011 (CUB) dataset~\cite{wah2011caltech} and the background images in the Places dataset~\cite{zhou2017places}.
There exist two attributes in the dataset; one is the type of a bird, \{waterbird, landbird\}, which is the target attribute, and the other is the background place, \{water, land\}.

The Colored-MNIST dataset~\cite{li2019repair, LfF, Rebias} is an extension of MNIST with the color attributes, where each digit is highly correlated to a certain color.
There are 60K training examples and 10K test images, where the ratio of bias-aligned samples\footnote{It denotes the samples that can be correctly classified by using the bias attribute (color).} is 95\%.
We follow the protocol employed in \cite{LfF} for the experiment.

\subsection{Implementation details}
For CelebA and Waterbirds, we use a ResNet-18 as our backbone network, which is pretrained on ImageNet. 
We train both the base and target models using the Adam optimizer with a learning rate of $1\times 10^{-4}$, a batch size of 256, and a weight decay rate of 0.01.
For the Colored MNIST dataset, we adopt a multi-layer perceptron with three hidden layers, each of which has 100 hidden units.
We also employ the same Adam optimizer with a learning rate of $1\times 10^{-3}$.
We train the models for 100 epochs for all experiments, and decay the learning rate using the cosine annealing~\cite{loshchilov2016sgdr}.

For clustering, we extract features from a separately trained base network with the standard classification loss and perform $k$-means clustering with $K=8$ in all experiments.
The cluster weight of the $k^\text{th}$ cluster, $\omega_k$, is updated by exponential moving average at each iteration with a momentum $m$ of 0.3.
All the results reported in our paper are obtained from the average of three runs.

\subsection{Evaluation protocol}
To evaluate the unbiased accuracy with an imbalanced evaluation set, we measure the accuracy of each group $g = ({a}_t$, ${a}_b)$, defined by a pair of target and bias attribute values.
We finally report the average accuracy of each group and worst-group accuracy among all groups as in \cite{LfF}.

\subsection{Results}
\label{sec:results}

We present our main results on the standard benchmarks including CelebA, WaterBirds, and Colored-MNIST.
In the rest of this section, our method is denoted by BPA, which stands for bias pseudo-attribute.

\vspace{-2mm}
\paragraph{CelebA}
\label{sec:celeba}

Before evaluating our frameworks, we first thoroughly analyze the CelebA dataset in terms of algorithmic bias among the attributes.
There are a total of 40 attributes in the CelebA dataset.
The bias attribute is fixed to \textit{gender}, and we analyze its relation to the attributes of the target candidates given by the rest of 39 attributes.
We accept the target attributes if the smallest group given by its combinations with the bias attribute in the test split contains at least 10 examples for statistical stability\footnote{The removed target attributes are \textit{5 o'clock shadow}, \textit{bald}, \textit{rosy cheeks}, \textit{sideburns}, \textit{goatee}, \textit{mustache}, and \textit{wearing necktie}.}.
We suppose that the spurious correlation exists between target and bias attributes when a baseline model gives a large performance gap between its overall accuracy and unbiased accuracy (\eg, $>$5\% points).
We found that 26 out of 32 attributes have spurious correlation to \textit{gender}, and report the results for the attributes.
See our supplementary files for more detailed analysis.

Table~\ref{tab:celeba} presents the experimental results of the proposed algorithm on the CelebA dataset, in comparison to the existing methods as well as the baseline model.
Our model significantly outperforms the baseline and LfF~\cite{LfF} for all target attributes in terms of both unbiased and worst-group accuracies.
Note that our model is almost competitive to a supervised approach, Group DRO~\cite{GroupDRO}, without the explicit bias information. 
On the other hand, we observe that training the model with LfF deteriorates performance even compared to the baseline.
This is because it fixes the feature extractor and only trains its classification layer at the end\footnote{\url{https://github.com/alinlab/LfF}}.
To conduct a meaningful comparison with the stable version of LfF, we first train the baseline model used in our experiment for 100 epochs and then fine-tune the classification layer only using the LfF algorithm; this revised version is referred to as LfF$^*$ in the rest of this section.
Although the performance of LfF$^*$ is stable, the improvement by debiasing is still limited compared to Group DRO and our approach.
Additional experimental results for other bias attributes are provided in our supplementary documents.

\begin{table*}[!ht]
\begin{center}
\caption{Unbiased and worst-group accuracies on the CelebA dataset with the target attributes, where the algorithmic bias does not exist.}
\vspace{-0.1cm}
\label{tab:celeba_unbiased}
 \scalebox{0.85}{
\setlength\tabcolsep{6pt} \hspace{-0.2cm}
\begin{tabular}{c|c|ccc|c|c||ccc|c|c}
\toprule
\multicolumn{2}{c|}{}&\multicolumn{5}{c||}{Unbiased accuracy (\%)}  &\multicolumn{5}{c}{Worst-group accuracy (\%)} \vspace{0.1cm} \\
\cdashline{3-12}
&&\multicolumn{4}{c|}{Unsupervised} & Supervised & \multicolumn{4}{c|}{Unsupervised} & Supervised \\
    Target & Bias & Base & LfF & LfF$^*$ & BPA (ours) & Group DRO & Base & LfF & LfF$^*$ & BPA (ours) & Group DRO\\
\hline
Attractive& Gender &76.05&30.18&75.97&{\underline{77.90}}&\bf{78.35}		&63.61&6.09&64.78&{\underline{65.20}}&{\bf66.30}\\
Smiling& Gender &91.66&74.62&91.20&\bf{92.08}&{\underline{91.64}}		&88.49&60.09&{\underline{88.65}}&\bf{90.06}&88.48\\
Mouth Open& Gender &93.10&81.85&92.96&{\underline{93.45}}&\bf{93.64}		&91.52&66.92&{\underline{92.44}}&\bf{92.27}&91.69\\
High Cheekbones& Gender &83.44&48.40&83.70&\bf{84.93}&{\underline{84.52}}	&70.49&7.92&73.56&\bf{78.56}&{\underline{78.37}}\\
Eyeglasses& Gender &98.20&85.47&{98.38}&{\underline{98.39}}&\bf{98.65}		&96.24&76.89&96.85&{\underline{97.22}}&\bf{97.64}\\
Black Hair& Gender &84.92&61.00&85.19&{\underline{86.57}}&\bf{86.76}		&75.47&22.04&75.69&\bf{81.28}&{\underline{80.67}}\\
\hline
    \bf{Average} & Gender & 87.90 & 63.59&  87.92 &  {\underline{88.89}}  & \bf{88.93}  		& 80.97 &39.51 &  82.29 &  \bf{84.10}  & {\underline{83.86}}  \\
\bottomrule
\end{tabular}
 }
\end{center}
\vspace{-0.2cm}
\end{table*}

\vspace{-2mm}
\paragraph{Waterbirds}
We also evaluate our model on the Waterbirds dataset and present the results in Table~\ref{tab:waterbirds}.
As in the CelebA dataset, our model achieves the best accuracies among the unsupervised methods in terms of both the unbiased and worst-group accuracies, and presents comparable results to the supervised method~\cite{GroupDRO}.
Our successful results on Waterbirds imply that the proposed method is robust to small-scale datasets as well.

\vspace{-2mm}
\paragraph{Colored-MNIST}
Table~\ref{tab:mnist} shows that our model achieves consistent accuracy in the digit classification with color bias.
Besides, the color classification performance, where the algorithmic bias does not exist, turns out to be also competitive to the baseline, while the supervised approach is not good at this setting.

\begin{table*}[t]
\begin{center}
\caption{Unbiased accuracy~(\%) with multiple bias attributes.
For each target, our model only requires a single model, while Group DRO~\cite{GroupDRO} should train separate models depending on the bias set.}
\vspace{-0.cm}
\label{tab:multi_bias}
\scalebox{0.85}{
\setlength\tabcolsep{5pt} \hspace{-0.2cm}
\hspace{-0.2cm}
\begin{tabular}{l|ccc|c||ccc|c}
\toprule
\multicolumn{1}{r|}{Target} & \multicolumn{4}{c||}{Blond Hair} & \multicolumn{4}{c}{Blurry} \\
\hline
&\multicolumn{3}{c|}{Unsupervised} & Supervised &\multicolumn{3}{c|}{Unsupervised} & Supervised \\
Biases & Baseline & LfF$^*$ & BPA (ours) & Group DRO & Baseline & LfF$^*$ & BPA (ours) & Group DRO \\
\hline
Gender & 80.42 &84.89 &\underline{90.18} & \bf{91.39} & 73.05 & 76.70 & \bf{88.93} & \underline{87.05}\\
\hline
Gender, Heavy Makeup 	& 83.64 & \underline{88.82}& \bf{91.90} & 81.09 	&75.37 & \underline{79.55} & \bf{89.09} & 72.17\\
Gender, Wearing Lipstick	& 80.34 & 84.13& \bf{91.63} & \underline{85.93} 	&79.88 & \underline{83.21} & \bf{89.79} & 79.88\\
Gender, Young			& 78.39 & 81.21 & \bf{89.05} & \underline{87.96}  	&72.97 & 77.77 & \bf{88.66} & \underline{85.39}\\
Gender, No Beard 		& 79.50 & 82.51 & \bf{89.92} & \underline{85.01} 	&78.91 & 79.84 & \bf{84.06} & \underline{81.07}\\
Gender, Wearing Necklace &79.25& 81.03 & \bf{92.62} & \underline{92.26} 		&71.80 & 78.07 & \bf{89.60} & \underline{85.33}\\
Gender, Big Nose 		& 81.18& 84.10& \underline{90.58} & \bf{90.83} 		&71.89 &	77.11 & \bf{88.57} & \underline{87.11}\\
Gender, Smiling 		& 79.75 & 82.91 & \underline{89.85} & \bf{91.73} 		& 73.31 & 78.04 & \bf{89.32} &\underline{87.87} \\
\hline
\bf{Average} & 80.29~{\footnotesize{$\pm$1.71}}& 83.53~{\footnotesize{$\pm$2.64}}& \bf{90.79}~{\footnotesize{$\pm$1.29}}& \underline{87.83}~{\footnotesize{$\pm$4.10}} 
&74.88~{\footnotesize{$\pm$3.32}}& 79.08~{\footnotesize{$\pm$2.06}}& \bf{88.44}~{\footnotesize{$\pm$1.98}}& \underline{82.69}~{\footnotesize{$\pm$5.50}} \\
\bottomrule
\end{tabular}
}
\end{center}
\vspace{-0.3cm}
\end{table*}

\subsection{Analysis}

\paragraph{Results with no algorithmic bias}
We test our algorithm on unbiased datasets to make sure that it is dependable on the cases without algorithmic bias.
The unbiased setting is defined by the configuration that a baseline model involves a marginal difference {between its overall accuracy and unbiased accuracy} (\eg, $<5$\% points).
Similar to Table~\ref{tab:celeba}, we identify a subset of target attributes in CelebA, which is not spuriously correlated to gender; there exist 6 out of 32 attributes.
Table~\ref{tab:celeba_unbiased} illustrates the results of the 6 target attributes, where the accuracy of our approach is most consistent among the 4 methods.
This implies that our framework can be incorporated into the existing recognition models directly, without knowing the presence of dataset bias. 
Note that color classification with digit bias on Colored-MNIST or background place classification with object bias on Waterbirds are also qualified as unbiased settings, where our model gives consistent results.


\vspace{-2mm}
\paragraph{Multiple bias attributes}
Thanks to the unsupervised nature of our method, we can simply evaluate our model on multi-bias scenarios, where multiple bias attributes exist in the dataset, without modification.
Table~\ref{tab:multi_bias} presents the unbiased results with multiple bias attributes using our method and Group DRO~\cite{GroupDRO}, where we additionally report the average and standard deviation over all bias sets to compare the overall effectiveness and robustness.
We also present the results of LfF$^*$, a variant of LfF~\cite{LfF}, introduced in Table~\ref{tab:celeba}.
When trained on multiple bias attributes, the accuracy of Group DRO is sensitive to bias sets while our method achieves stable and superior results for a variety of sets.
Note also that our model is applicable to any bias sets without additional fine-tuning while a supervised method should separately train their model for each set.

\vspace{-2mm}
\paragraph{Ablative results on importance weighting}

\begin{table}[t]
\begin{center}
\caption{Ablation results on our importance weighting scheme on CelebA with \textit{blond hair} and \textit{gender} for target and bias attributes, respectively, in terms of unbiased and worst-group accuracies (\%).}
\vspace{-0.1cm}
\label{tab:ablation_weighting}
\scalebox{0.85}{
\setlength\tabcolsep{12pt}
\begin{tabular}{cc|cc}
\toprule
Scale & Loss  & Unbiased Acc. & Worst-group Acc. \\
\hline
& &  80.42 & 41.02\\
 &\checkmark & 83.86 & 57.44 \\
 \checkmark& & 89.08 & 76.55\\
\checkmark &\checkmark &\bf{90.18} &\bf{82.54}\\
\bottomrule
\end{tabular}
}
\end{center}
\vspace{-2mm}
\end{table}

We perform the ablative experiments on the CelebA dataset to analyze the effectiveness of our cluster weighting strategy.
Table~\ref{tab:ablation_weighting} presents the results when only one of the scale $N_k$ and the average loss ${\mathbb{E}_{(\mathbf{x}, y) \sim {P_k}}[\ell( (\mathbf{x}, y); {\theta})]}$, respectively, are taken into account to compute $\omega_k$ for the $k^\text{th}$ cluster in Eq.~\eqref{eq:importance}.
Note that our ablative model with the loss factor only is closely related to~\cite{sohoni2020no}.
Table~\ref{tab:ablation_weighting} also shows that combining both terms plays a crucial role for learning debiased representations while the scale factor is clearly more important.

\begin{table}[t]
\begin{center}
\caption{Unbiased accuracy ($\%$) with bias-unspecified settings.
The results are the average of a set of unbiased accuracies, each of which adopts one of the  25 unspecified attributes jointly with the specified bias attribute, \textit{gender}, as the bias attributes to define groups.
}
\label{tab:celeba_unspecified}
\vspace{-0.1cm}
\scalebox{0.85}{
\setlength\tabcolsep{9pt} \hspace{-0.2cm}
\begin{tabular}{c|cc|c}
\toprule
        Target & {Baseline} &  BPA (ours) & Group DRO \\
\hline
Blond Hair&79.13~\footnotesize{$\pm$~2.72}	& 90.16~\footnotesize{$\pm$~3.19}  & {\bf{90.82}}~\footnotesize{$\pm$~2.76} \\
Heavy Makeup &  70.26~\footnotesize{$\pm$~3.84}   & {\bf{73.52}}~\footnotesize{$\pm$~3.86}  &  71.57~\footnotesize{$\pm$~4.33}   		\\
Young  &   77.56~\footnotesize{$\pm$~1.80}    &  {\bf{81.31}}~\footnotesize{$\pm$~2.31}  &   80.56~\footnotesize{$\pm$~2.09}  	\\
Double Chin &   62.56~\footnotesize{$\pm$~2.22}   & {\bf{81.71}}~\footnotesize{$\pm$~3.61}  &  78.46~\footnotesize{$\pm$~3.37}  	\\
Chubby &   67.80~\footnotesize{$\pm$~2.77}   & {\bf{82.36}}~\footnotesize{$\pm$~3.55}  &   79.91~\footnotesize{$\pm$~3.43}  		\\
Wearing Hat &     90.80~\footnotesize{$\pm$~4.01}    &  {\bf{95.11}}~\footnotesize{$\pm$~3.10}  &    94.71~\footnotesize{$\pm$~3.76} 		\\
Oval Face &     61.77~\footnotesize{$\pm$~1.80}    & {\bf{66.63}}~\footnotesize{$\pm$~1.63}  &    65.35~\footnotesize{$\pm$~1.45} 	\\
Pointy Nose &  63.96~\footnotesize{$\pm$1.42}   &  {\bf{70.53}}~\footnotesize{$\pm$1.52}  &  70.16~\footnotesize{$\pm$1.45} 	\\
\bottomrule
\end{tabular}
}
\end{center}
\vspace{-0.3cm}
\end{table}

\vspace{-2mm}
\paragraph{{Unspecified group shifts}}
To verify the robustness in another realistic scenario, we test unspecified group shifts, where the group information at test time is not fully provided during training.
The bias attribute specified during training, which is exploited by group DRO, is fixed to \textit{gender}.
To evaluate the performance in this setting, we measure a set of unbiased accuracies corresponding to the combinations of the specified bias attribute, \textit{gender}, and each of 25 unspecified bias attributes except the target attribute\footnote{There exist 26 (out of 39) valid attributes as described in Section~\ref{sec:celeba}.}.
Note that the unspecified bias attributes are not introduced during training but used to define groups at test time.
Table~\ref{tab:celeba_unspecified} clearly shows that our model outperforms Group DRO in the bias-unspecified setting, where we only report the average and standard deviation of the set of unbiased accuracies due to space limitation.
This implies that, although Group DRO handles group shifts well within the simplex of the specified group distributions, it suffers from worst-case generalization for unspecified group shifts.
Note that the proposed approach is free from the issue because it does not use any information about bias for training.

\begin{figure}
	\begin{center}
        \includegraphics[width=0.95\linewidth]{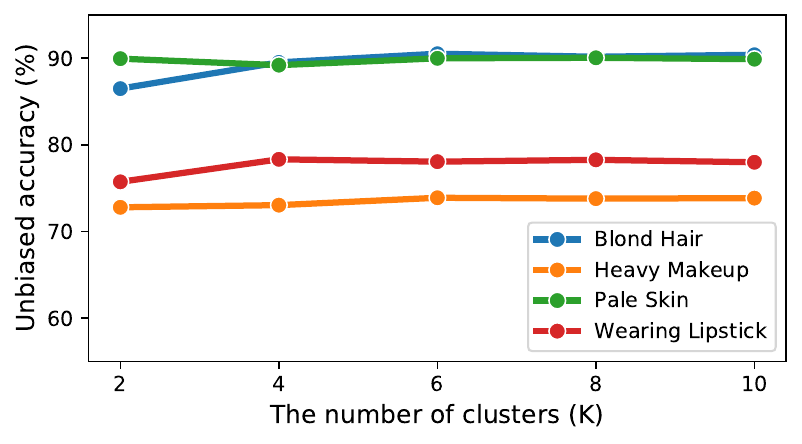}
        \vspace{-0.5cm}
	\end{center}
	\caption{Sensitivity analysis on the number of clusters in our framework on the CelebA dataset.
	}
	\label{fig:abl_k}
	\vspace{0.1cm}    
\end{figure}

\vspace{-2mm}
\paragraph{Sensitivity analysis on the number of clusters}
We conduct ablation study on the number of groups for clustering on the feature embedding space to obtain bias pseudo-attributes on the CelebA dataset.
We set \textit{gender} as the bias attribute and evaluate the unbiased accuracies for several target attributes.
Figure~\ref{fig:abl_k} presents that the results are stable over a wide range of the number of clusters and the accuracy is saturated when $K \geq 4$.

\begin{figure}
	\begin{center}
        \includegraphics[width=0.97\linewidth]{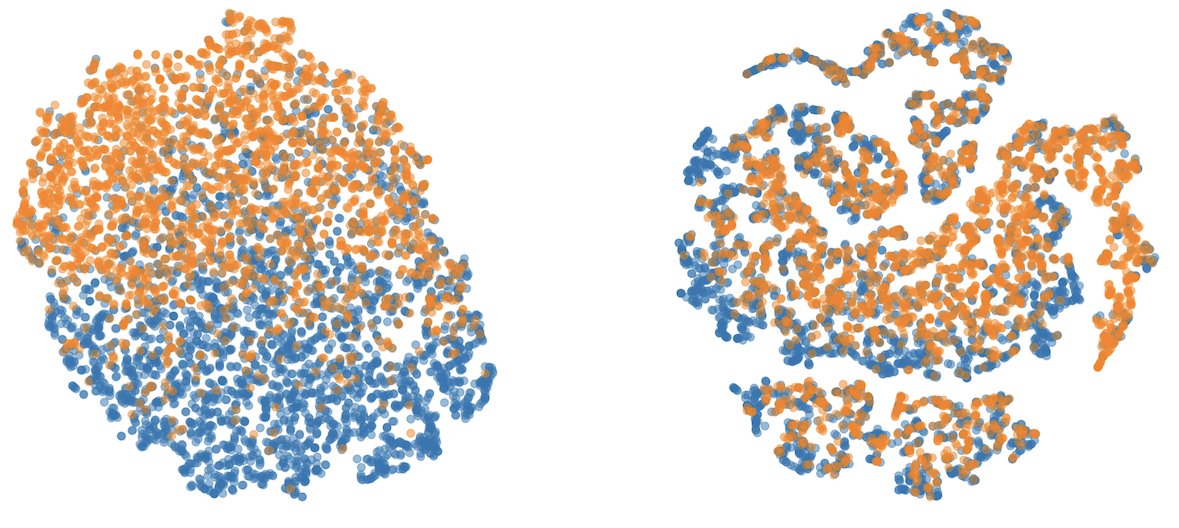}
        \vspace{-0.5cm}
	\end{center}
	\caption{
	The t-SNE plots of feature embeddings using baseline (left) and ours (right) trained to classify \textit{hair color}.
	We visualize the distribution of samples that have the same target value (\textit{blond hair = false}).
	Blue and orange colors denote different gender values, female and male, respectively.
	Our framework helps mix samples which have the same target but different bias attribute values.
	}
	\label{fig:feature_emb}
	\vspace{-0.0cm}    
\end{figure}

\vspace{-2mm}
\paragraph{Feature visualization} 
Figure~\ref{fig:feature_emb} illustrates the t-SNE results of instance embeddings for the baseline model (left) and ours (right) on the CelebA dataset for the \textit{blond hair} attribute classification, where we visualize only negative (\textit{blond hair = false}) examples for an effective visualization.
Blue and orange colors indicate values---female and male, respectively---of the bias attribute, \textit{gender}.
We observe that our model helps to mix the examples in different groups within the same class, which is desirable for debiasing.

\section{Conclusion}
\label{sec:conclusion}
We presented a generic unsupervised debiasing framework using pseudo-attribute.
We observed that the examples sampled from the same groups are closely located in the feature embedding space.
Based on our empirical observation, we claim that it is possible to identify the pseudo-attributes by taking advantage of the embedding results even without the attribute supervision.
Inspired by this fact, we introduced a novel cluster-based weighting strategy for learning debiased representations.
We demonstrated the effectiveness of our method on multiple standard benchmarks, which is even as competitive as the supervised debiasing method, group DRO.
We also conducted a thorough analysis of our framework in many realistic scenarios, where our model provides substantial gains consistently.

\vspace{-2mm}
\paragraph{Potential societal impact and limitation}
Machine learning models typically focus on performance improvement unconditionally.
Hence, it is often exposed to the risk caused by dataset and/or algorithmic bias, which need to be carefully addressed for enhancing reliability and robust of models.
This research contributes to a positive societal impact from this point of view.
Although the proposed algorithm turns out to be effective for bias identification, there may be blind spots due to unexplored types of bias.
Therefore, we believe that the identification of hidden and unobservable biases without prior knowledge is a promising research direction.

\vspace{-2mm}
\paragraph{Acknowledgments}

This work was partly supported by the IITP grants [2021-0-02068, Artificial Intelligence Innovation Hub; 2021-0-01343, Artificial Intelligence Graduate School Program (Seoul National University)] and the National Research Foundation of Korea (NRF) grant [2022R1A2C3012210] funded by the Korean government (MSIT) and by Samsung Electronics Co., Ltd. [IO210917-08957-01].

{\small
\bibliographystyle{ieee_fullname}
\bibliography{egbib}
}


\onecolumn

\setcounter{section}{0}
\setcounter{table}{0}
\setcounter{figure}{0}
\renewcommand\thesection{\Alph{section}}
\renewcommand\thetable{\Alph{table}}
\renewcommand\thefigure{\Alph{figure}}

\vspace{-1cm}

\section{Full experimental results}

Table~\ref{tab:celeba_full} and \ref{tab:celeba_worst_full} present the full results of Table 1 in the main paper, including George~\cite{sohoni2020no} and a class weighting method.
George~\cite{sohoni2020no} is closely related to our ablative model with sample weighting based on its loss, which is shown in Table 6 of the main paper, while class weighting approach adjusts the weight of each example depending on the associated class scale (size) to mitigate the class imbalance issue.
We also report the gap between the overall accuracy and the unbiased accuracy of the baseline model to present the degree of algorithmic bias for each target attribute with \textit{gender} bias. 
Bold and underline fonts indicate the first and second place among the compared approaches, respectively.
The proposed approach achieves outstanding performance compared to all other unsupervised methods, and is even as competitive as the supervised counterpart~\cite{GroupDRO}.
Also, it is surprising that the class weighting method is superior to existing unsupervised debiasing methods including LfF~\cite{LfF} and George~\cite{sohoni2020no}.
We run all experimental three times and compute average accuracies and their standard deviations.

\vspace{0.2cm}
\begin{table*}[h]
\begin{center}
\caption{Unbiased accuracy ($\%$) in the presence of spurious correlations between target and bias attributes on the test split of the CelebA dataset.}
\vspace{-0.2cm}
\label{tab:celeba_full}
 \scalebox{0.83}{
 \setlength\tabcolsep{6pt} \hspace{-0.25cm}
\begin{tabular}{c|c|c|ccccc||c}
\toprule
&&&\multicolumn{5}{c||}{Unsupervised} & Supervised \\
    Target  &  Gap (\%p) &Overall & Baseline & LfF$^{\text{*}}$~\cite{LfF} & George~\cite{sohoni2020no} & Class weighting & \bf{Ours} & Group DRO~\cite{GroupDRO}\\
\hline
\hline
Blond Hair&-15.28&95.70&80.42~\footnotesize{$\pm$~0.51}&84.89~\footnotesize{$\pm$~0.14}&83.13~\footnotesize{$\pm$~1.86}&83.35~\footnotesize{$\pm$~0.85}&{\underline{90.18}~\footnotesize{$\pm$~0.23}}&{\bf91.39}~\footnotesize{$\pm$~0.27}\\
Heavy Makeup&-19.63&90.82&71.19~\footnotesize{$\pm$~0.37}&71.85~\footnotesize{$\pm$~0.17}&70.91~\footnotesize{$\pm$~0.77}&71.74~\footnotesize{$\pm$~0.83}&\bf73.78~\footnotesize{$\pm$~0.25}&{\underline{72.70}~\footnotesize{$\pm$~0.71}}\\
Pale Skin&-25.25&96.75&71.50~\footnotesize{$\pm$~1.60}&75.23~\footnotesize{$\pm$~0.74}&78.22~\footnotesize{$\pm$~3.75}&90.02~\footnotesize{$\pm$~0.56}&{\underline{90.06}~\footnotesize{$\pm$~0.75}}&\bf90.55~\footnotesize{$\pm$~0.84}\\
Wearing Lipstick&-18.70&92.60&73.90~\footnotesize{$\pm$~0.53}&73.84~\footnotesize{$\pm$~0.05}&78.05~\footnotesize{$\pm$~0.98}&72.89~\footnotesize{$\pm$~1.28}&\bf78.28~\footnotesize{$\pm$~0.88}&{\underline{78.26}~\footnotesize{$\pm$~2.73}}\\
Young&-9.30&87.49&78.19~\footnotesize{$\pm$~0.39}&79.58~\footnotesize{$\pm$~0.14}&80.79~\footnotesize{$\pm$~0.20}&82.13~\footnotesize{$\pm$~0.82}&{\underline{82.27}~\footnotesize{$\pm$~0.65}}&\bf82.40~\footnotesize{$\pm$~0.48}\\
Double Chin&-31.32&95.93&64.61~\footnotesize{$\pm$~0.82}&68.47~\footnotesize{$\pm$~0.22}&76.23~\footnotesize{$\pm$~0.11}&82.13~\footnotesize{$\pm$~1.43}&{\underline{82.92}~\footnotesize{$\pm$~0.54}}&\bf83.19~\footnotesize{$\pm$~1.11}\\
Chubby&-27.97&95.39&67.42~\footnotesize{$\pm$~0.95}&71.56~\footnotesize{$\pm$~0.52}&74.88~\footnotesize{$\pm$~1.91}&79.64~\footnotesize{$\pm$~0.56}&\bf83.88~\footnotesize{$\pm$~0.36}&{\underline{81.90}~\footnotesize{$\pm$~0.20}}\\
Wearing Hat&-5.57&99.10&93.53~\footnotesize{$\pm$~0.37}&94.81~\footnotesize{$\pm$~0.15}&95.72~\footnotesize{$\pm$~0.71}&96.16~\footnotesize{$\pm$~0.50}&{\underline{96.80}~\footnotesize{$\pm$~0.26}}&\bf96.84~\footnotesize{$\pm$~0.46}\\
Oval Face&-10.40&73.10&62.70~\footnotesize{$\pm$~0.62}&62.30~\footnotesize{$\pm$~0.21}&65.16~\footnotesize{$\pm$~0.23}&65.13~\footnotesize{$\pm$~1.05}&\bf67.18~\footnotesize{$\pm$~0.82}&{\underline{65.40}~\footnotesize{$\pm$~0.14}}\\
Pointy Nose&-11.81&73.91&62.10~\footnotesize{$\pm$~0.74}&63.83~\footnotesize{$\pm$~0.28}&61.68~\footnotesize{$\pm$~1.59}&66.82~\footnotesize{$\pm$~2.76}&{\underline{68.90}~\footnotesize{$\pm$~0.90}}&\bf70.71~\footnotesize{$\pm$~0.28}\\
Straight Hair&-12.24&82.52&70.28~\footnotesize{$\pm$~1.06}&72.84~\footnotesize{$\pm$~0.12}&{\underline{77.80}~\footnotesize{$\pm$~0.19}}&77.46~\footnotesize{$\pm$~0.70}&\bf79.18~\footnotesize{$\pm$~0.38}&{77.04~\footnotesize{$\pm$~0.70}}\\
Blurry&-22.98&96.03&73.05~\footnotesize{$\pm$~1.28}&77.52~\footnotesize{$\pm$~0.45}&81.28~\footnotesize{$\pm$~0.28}&{\underline{87.75}~\footnotesize{$\pm$~0.87}}&\bf88.93~\footnotesize{$\pm$~0.32}&{87.05~\footnotesize{$\pm$~0.90}}\\
Narrow Eyes&-23.29&86.47&63.18~\footnotesize{$\pm$~1.05}&67.77~\footnotesize{$\pm$~0.08}&68.03~\footnotesize{$\pm$~0.11}&70.99~\footnotesize{$\pm$~0.60}&{\underline{76.39}~\footnotesize{$\pm$~0.64}}&\bf76.72~\footnotesize{$\pm$~1.98}\\
Arched Eyebrows&-12.09&81.81&69.72~\footnotesize{$\pm$~0.37}&71.87~\footnotesize{$\pm$~0.10}&73.25~\footnotesize{$\pm$~0.29}&{\underline{75.58}~\footnotesize{$\pm$~1.13}}&74.77~\footnotesize{$\pm$~0.69}&\bf78.30~\footnotesize{$\pm$~1.79}\\
Bags Under Eyes&-14.16&83.63&69.47~\footnotesize{$\pm$~0.57}&71.86~\footnotesize{$\pm$~0.05}&74.81~\footnotesize{$\pm$~0.38}&{\underline{76.36}~\footnotesize{$\pm$~1.05}}&\bf77.84~\footnotesize{$\pm$~1.14}&75.88~\footnotesize{$\pm$~1.18}\\
Bangs&-6.37&95.41&89.04~\footnotesize{$\pm$~0.47}&89.04~\footnotesize{$\pm$~0.50}&92.62~\footnotesize{$\pm$~0.12}&93.09~\footnotesize{$\pm$~0.29}&{\underline{93.94}~\footnotesize{$\pm$~0.57}}&\bf94.45~\footnotesize{$\pm$~0.17}\\
Big Lips&-8.99&69.86&60.87~\footnotesize{$\pm$~0.58}&62.15~\footnotesize{$\pm$~0.06}&{\underline{64.99}~\footnotesize{$\pm$~0.13}}&{63.74}~\footnotesize{$\pm$~0.56}&\bf66.50~\footnotesize{$\pm$~0.24}&63.70~\footnotesize{$\pm$~0.44}\\
No Beard&-22.73&95.84&73.11~\footnotesize{$\pm$~0.90}&73.13~\footnotesize{$\pm$~0.89}&{\underline{77.90}~\footnotesize{$\pm$~0.20}}&77.83~\footnotesize{$\pm$~2.29}&\bf79.58~\footnotesize{$\pm$~0.14}&{77.86~\footnotesize{$\pm$~1.35}}\\
Receding Hairline&-23.31&93.03&69.72~\footnotesize{$\pm$~0.78}&74.58~\footnotesize{$\pm$~0.21}&78.86~\footnotesize{$\pm$~0.40}&82.97~\footnotesize{$\pm$~0.97}&{\underline{84.95}~\footnotesize{$\pm$~0.49}}&\bf85.15~\footnotesize{$\pm$~1.31}\\
Wavy Hair&-9.19&82.29&73.10~\footnotesize{$\pm$~0.56}&74.53~\footnotesize{$\pm$~0.17}&77.39~\footnotesize{$\pm$~0.15}&76.50~\footnotesize{$\pm$~0.65}&\bf79.89~\footnotesize{$\pm$~0.71}&{\underline{79.65}~\footnotesize{$\pm$~0.63}}\\
Wearing Earrings&-17.18&89.35&72.17~\footnotesize{$\pm$~0.91}&74.17~\footnotesize{$\pm$~0.33}&80.65~\footnotesize{$\pm$~0.04}&78.65~\footnotesize{$\pm$~0.28}&\bf84.57~\footnotesize{$\pm$~0.69}&{\underline{83.50}~\footnotesize{$\pm$~0.63}}\\
Wearing Necklace&-30.73&85.77&55.04~\footnotesize{$\pm$~0.59}&57.21~\footnotesize{$\pm$~0.76}&58.79~\footnotesize{$\pm$~0.10}&{\underline{67.05}~\footnotesize{$\pm$~1.37}}&\bf68.96~\footnotesize{$\pm$~0.12}&62.89~\footnotesize{$\pm$~3.69}\\
Big Nose&-14.74&82.44&67.70~\footnotesize{$\pm$~1.11}&69.75~\footnotesize{$\pm$~0.03}&71.85~\footnotesize{$\pm$~0.18}&70.52~\footnotesize{$\pm$~1.02}&\bf74.21~\footnotesize{$\pm$~0.43}&{\underline{73.73}~\footnotesize{$\pm$~0.27}}\\
Brown Hair&-8.88&86.95&78.07~\footnotesize{$\pm$~0.87}&78.93~\footnotesize{$\pm$~1.24}&83.07~\footnotesize{$\pm$~0.07}&83.12~\footnotesize{$\pm$~0.38}&{\underline{83.83}~\footnotesize{$\pm$~0.66}}&\bf84.87~\footnotesize{$\pm$~0.07}\\
Bushy Eyebrows&-17.02&91.44&74.42~\footnotesize{$\pm$~0.91}&75.20~\footnotesize{$\pm$~0.34}&80.99~\footnotesize{$\pm$~0.32}&82.73~\footnotesize{$\pm$~1.21}&{\underline{85.02}~\footnotesize{$\pm$~0.02}}&\bf85.43~\footnotesize{$\pm$~0.19}\\
Gray Hair&-20.54&98.01&77.47~\footnotesize{$\pm$~0.67}&80.09~\footnotesize{$\pm$~0.21}&86.10~\footnotesize{$\pm$~1.18}&90.12~\footnotesize{$\pm$~1.12}&{\underline{91.80}~\footnotesize{$\pm$~0.22}}&\bf92.52~\footnotesize{$\pm$~0.14}\\
\hline
\bf{Average} & -16.91 & 88.52 & 71.61 & 73.73 & 76.66 & 78.63 & \bf{80.93} & {\underline{80.46}} \\
\bottomrule
\end{tabular}
 }
\end{center}
\end{table*}

\begin{table*}[t]
\begin{center}
\caption{Worst-group accuracy ($\%$) in the presence of spurious correlation between target and bias attributes on the test split of the CelebA dataset.}
\label{tab:celeba_worst_full}
 \scalebox{0.83}{
 \setlength\tabcolsep{6pt} \hspace{-0.25cm}
\begin{tabular}{c|c|c|ccccc||c}
\toprule
&&&\multicolumn{5}{c||}{Unsupervised} & Supervised \\
    Target  &  Gap (\%p) &Overall & Baseline & LfF$^{\text{*}}$~\cite{LfF} & George~\cite{sohoni2020no} & Class weighting & \bf{Ours} & Group DRO~\cite{GroupDRO}\\
\hline
\hline
Blond Hair&-54.68&95.70&41.02~\footnotesize{$\pm$~1.96}&57.96~\footnotesize{$\pm$~2.00}&65.45~\footnotesize{$\pm$~15.52}&53.58~\footnotesize{$\pm$~3.10}&{\underline{82.54}~\footnotesize{$\pm$~1.22}}&\bf87.86~\footnotesize{$\pm$~0.10}\\
Heavy Makeup&-73.47&90.82&17.35~\footnotesize{$\pm$~4.60}&23.87~\footnotesize{$\pm$~2.79}&9.09~\footnotesize{$\pm$~1.24}&{\underline{28.86}~\footnotesize{$\pm$~11.91}}&\bf39.84~\footnotesize{$\pm$~2.28}&21.36~\footnotesize{$\pm$~1.36}\\
Pale Skin&-60.11&96.75&36.64~\footnotesize{$\pm$~3.53}&43.26~\footnotesize{$\pm$~1.40}&62.03~\footnotesize{$\pm$~16.50}&85.42~\footnotesize{$\pm$~1.70}&\bf88.60~\footnotesize{$\pm$~1.48}&{\underline{87.68}~\footnotesize{$\pm$~2.37}}\\
Wearing Lipstick&-61.22&92.60&31.38~\footnotesize{$\pm$~4.27}&31.92~\footnotesize{$\pm$~0.02}&51.04~\footnotesize{$\pm$~2.59}&27.68~\footnotesize{$\pm$~3.45}&\bf46.52~\footnotesize{$\pm$~1.62}&{\underline{46.08}~\footnotesize{$\pm$~5.57}}\\
Young&-34.70&87.49&52.79~\footnotesize{$\pm$~1.45}&57.79~\footnotesize{$\pm$~0.84}&65.12~\footnotesize{$\pm$~0.88}&71.43~\footnotesize{$\pm$~1.75}&{\underline{74.33}~\footnotesize{$\pm$~0.70}}&\bf76.29~\footnotesize{$\pm$~1.96}\\
Double Chin&-74.60&95.93&21.33~\footnotesize{$\pm$~2.24}&28.24~\footnotesize{$\pm$~0.46}&50.00~\footnotesize{$\pm$~0.41}&62.43~\footnotesize{$\pm$~4.71}&{\underline{67.78}~\footnotesize{$\pm$~0.91}}&\bf72.94~\footnotesize{$\pm$~1.14}\\
Chubby&-71.09&95.39&24.30~\footnotesize{$\pm$~3.73}&34.09~\footnotesize{$\pm$~0.90}&58.01~\footnotesize{$\pm$~11.04}&52.76~\footnotesize{$\pm$~2.59}&{\underline{72.32}~\footnotesize{$\pm$~0.93}}&\bf72.64~\footnotesize{$\pm$~1.70}\\
Wearing Hat&-13.98&99.10&85.12~\footnotesize{$\pm$~0.31}&88.31~\footnotesize{$\pm$~0.12}&92.93~\footnotesize{$\pm$~0.76}&93.61~\footnotesize{$\pm$~0.32}&\bf94.94~\footnotesize{$\pm$~0.19}&{\underline{94.67}~\footnotesize{$\pm$~0.41}}\\
Oval Face&-43.95&73.10&29.15~\footnotesize{$\pm$~2.76}&36.00~\footnotesize{$\pm$~1.46}&38.01~\footnotesize{$\pm$~2.63}&43.52~\footnotesize{$\pm$~6.37}&{\underline{55.78}~\footnotesize{$\pm$~0.94}}&\bf56.84~\footnotesize{$\pm$~1.83}\\
Pointy Nose&-48.11&73.91&25.80~\footnotesize{$\pm$~4.03}&38.04~\footnotesize{$\pm$~1.49}&22.63~\footnotesize{$\pm$~3.67}&47.46~\footnotesize{$\pm$~3.75}&{\underline{52.48}~\footnotesize{$\pm$~0.52}}&\bf63.76~\footnotesize{$\pm$~2.80}\\
Straight Hair&-34.70&82.52&47.82~\footnotesize{$\pm$~6.75}&58.53~\footnotesize{$\pm$~1.61}&69.23~\footnotesize{$\pm$~1.24}&{\underline{68.97}~\footnotesize{$\pm$~1.15}}&\bf72.09~\footnotesize{$\pm$~0.76}&66.10~\footnotesize{$\pm$~3.56}\\
Blurry&-50.35&96.03&45.68~\footnotesize{$\pm$~3.98}&52.35~\footnotesize{$\pm$~1.18}&62.23~\footnotesize{$\pm$~1.58}&{\underline{82.30}~\footnotesize{$\pm$~3.05}}&\bf84.10~\footnotesize{$\pm$~0.73}&{82.06~\footnotesize{$\pm$~2.27}}\\
Narrow Eyes&-59.46&86.47&27.01~\footnotesize{$\pm$~1.30}&38.53~\footnotesize{$\pm$~0.44}&35.16~\footnotesize{$\pm$~1.14}&52.62~\footnotesize{$\pm$~4.11}&\bf73.24~\footnotesize{$\pm$~0.88}&{\underline{71.47}~\footnotesize{$\pm$~3.72}}\\
Arched Eyebrows&-47.05&81.81&34.76~\footnotesize{$\pm$~1.86}&44.97~\footnotesize{$\pm$~0.46}&45.64~\footnotesize{$\pm$~1.21}&52.94\footnotesize{$\pm$~5.28}&{\underline{54.36}~\footnotesize{$\pm$~1.37}}&\bf69.44~\footnotesize{$\pm$~5.44}\\
Bags Under Eyes&-41.98&83.63&41.65~\footnotesize{$\pm$~1.01}&49.10~\footnotesize{$\pm$~0.49}&56.28~\footnotesize{$\pm$~2.11}&59.77~\footnotesize{$\pm$~8.13}&{\underline{62.55}~\footnotesize{$\pm$~0.90}}&\bf63.34~\footnotesize{$\pm$~3.02}\\
Bangs&-18.50&95.41&76.91~\footnotesize{$\pm$~3.27}&82.37~\footnotesize{$\pm$~0.52}&85.90~\footnotesize{$\pm$~0.24}&87.91~\footnotesize{$\pm$~1.80}&\bf92.21~\footnotesize{$\pm$~1.24}&{\underline{92.12}~\footnotesize{$\pm$~1.03}}\\
Big Lips&-39.01&69.86&30.85~\footnotesize{$\pm$~0.62}&38.54~\footnotesize{$\pm$~0.18}&44.51~\footnotesize{$\pm$~0.83}&43.16~\footnotesize{$\pm$~5.62}&\bf56.99~\footnotesize{$\pm$~3.05}&{\underline{47.55}~\footnotesize{$\pm$~1.03}}\\
No Beard&-82.54&95.84&13.30~\footnotesize{$\pm$~3.87}&20.00~\footnotesize{$\pm$~0.00}&33.33~\footnotesize{$\pm$~5.77}&30.00~\footnotesize{$\pm$~10.00}&\bf40.00~\footnotesize{$\pm$~0.00}&{\underline{36.70}~\footnotesize{$\pm$~5.10}}\\
Receding Hairline&-57.34&93.03&35.69~\footnotesize{$\pm$~0.35}&45.53~\footnotesize{$\pm$~0.55}&57.30~\footnotesize{$\pm$~0.90}&72.14~\footnotesize{$\pm$~2.56}&\bf79.12~\footnotesize{$\pm$~1.91}&\bf79.12~\footnotesize{$\pm$~2.11}\\
Wavy Hair&-44.28&82.29&38.01~\footnotesize{$\pm$~0.85}&45.24~\footnotesize{$\pm$~0.83}&53.17~\footnotesize{$\pm$~0.43}&49.69~\footnotesize{$\pm$~4.65}&{\underline{65.74}~\footnotesize{$\pm$~1.13}}&\bf66.79~\footnotesize{$\pm$~1.62}\\
Wearing Earrings&-63.09&89.35&26.26~\footnotesize{$\pm$~4.14}&32.95~\footnotesize{$\pm$~1.31}&52.74~\footnotesize{$\pm$~1.10}&47.18~\footnotesize{$\pm$~4.08}&{\underline{72.81}~\footnotesize{$\pm$~1.50}}&\bf75.24~\footnotesize{$\pm$~2.10}\\
Wearing Necklace&-83.05&85.77&2.72~\footnotesize{$\pm$~0.83}&6.67~\footnotesize{$\pm$~2.07}&13.82~\footnotesize{$\pm$~0.41}&{\underline{30.36}~\footnotesize{$\pm$~3.36}}&\bf41.93~\footnotesize{$\pm$~2.47}&24.34~\footnotesize{$\pm$~7.81}\\
Big Nose&-49.25&82.44&33.19~\footnotesize{$\pm$~3.97}&45.30~\footnotesize{$\pm$~0.50}&46.22~\footnotesize{$\pm$~0.41}&49.56~\footnotesize{$\pm$~4.79}&{\underline{63.00}~\footnotesize{$\pm$~4.27}}&\bf65.08~\footnotesize{$\pm$~1.17}\\
Brown Hair&-27.37&86.95&59.58~\footnotesize{$\pm$~2.55}&60.68~\footnotesize{$\pm$~3.62}&73.20~\footnotesize{$\pm$~0.88}&70.91~\footnotesize{$\pm$~3.09}&{\underline{71.50}~\footnotesize{$\pm$~0.97}}&\bf78.92~\footnotesize{$\pm$~1.61}\\
Bushy Eyebrows&-54.30&91.44&37.14~\footnotesize{$\pm$~2.54}&52.67~\footnotesize{$\pm$~3.14}&56.08~\footnotesize{$\pm$~0.97}&66.92~\footnotesize{$\pm$~6.98}&{\underline{74.08}~\footnotesize{$\pm$~0.75}}&\bf81.56~\footnotesize{$\pm$~3.24}\\
Gray Hair&-55.52&98.01&42.49~\footnotesize{$\pm$~1.86}&48.46~\footnotesize{$\pm$~1.09}&67.23~\footnotesize{$\pm$~2.75}&80.00~\footnotesize{$\pm$~3.78}&{\underline{83.03}~\footnotesize{$\pm$~1.37}}&\bf88.55~\footnotesize{$\pm$~1.85}\\
\hline
\bf{Average} & 51.68 & 88.79 & 36.84 & 44.67 & 50.39 & 58.00 & {\underline{67.76}} & \bf{68.02} \\
\bottomrule
\end{tabular}
 }
\end{center}
\end{table*}

\clearpage

\section{Additional Analysis}

\begin{figure}
	\begin{center}
	\includegraphics[width=0.9\linewidth]{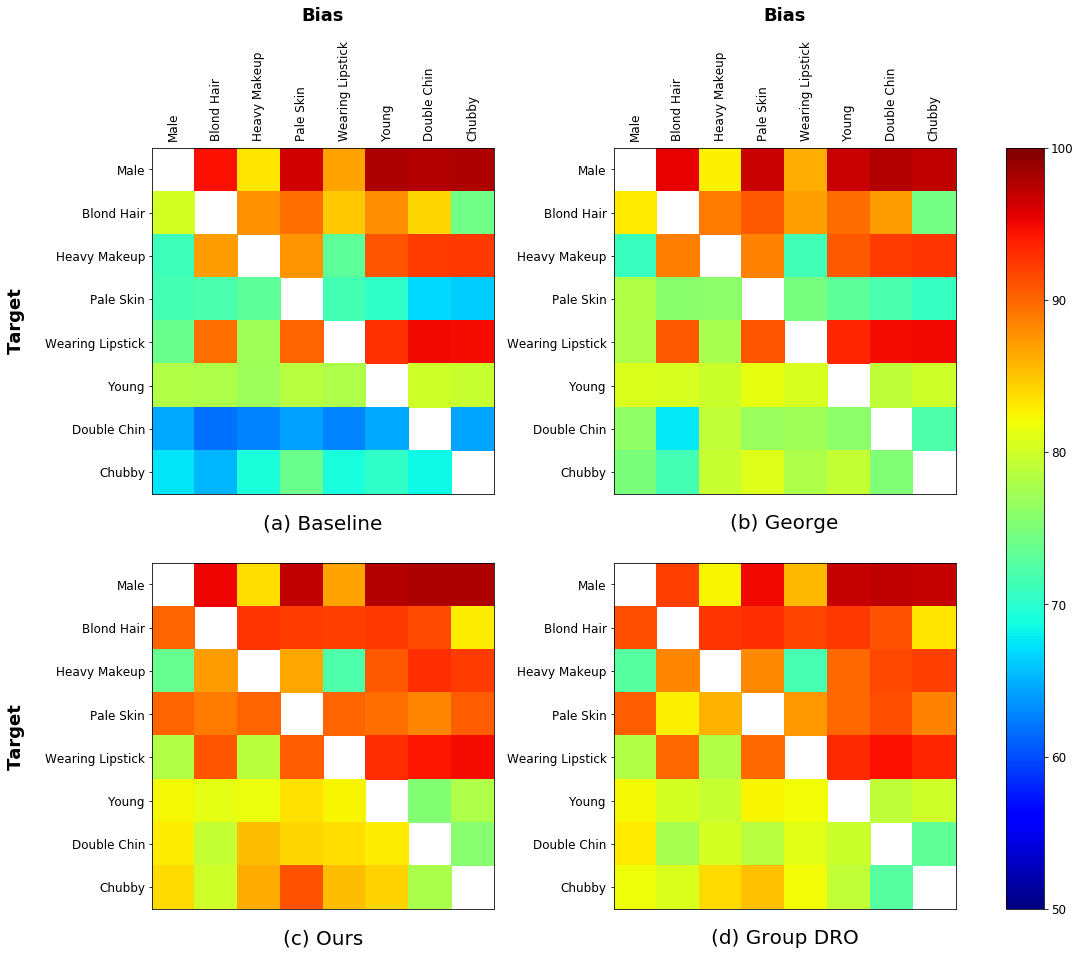}
	\end{center}
	\caption{
	Heatmap of unbiased accuracy (\%) with 4 different methods.
	Unlike previous tables, we evaluate our model with various bias attributes, in addition to \textit{Male} (gender), on the CelebA dataset.
	To be specific, we select 8 attributes and evaluate unbiased accuracies with all possible (target, bias) pairs among the attributes.
	For each figure, the columns and rows denote bias and target attributes, respectively.
	Our approach substantially improves unbiased accuracies for various bias attributes consistently.
	}
	\label{fig:heatmap}
\end{figure}

\paragraph{Unbiased results with various bias attributes}
To make our study more comprehensive, we also evaluate our model with various bias attributes, in addition to \textit{Male} (gender), on the CelebA dataset.
Specifically, we select 8 attributes\footnote{The selected attributes are male, blond hair, heavy makeup, pale skin, wearing lipstick, young, double chin and chubby.} and test our model with all possible (target, bias) pairs among the attributes.
Figure~\ref{fig:heatmap} visualizes the experimental results with different methods, including baseline, George~\cite{sohoni2020no}, group DRO~\cite{GroupDRO} and our approach, in terms of unbiased accuracy (\%).
The columns and rows denote bias and target attributes, respectively.
As shown in the figure, our model improves unbiased accuracies substantially for various bias attributes, which outperforms baseline and George~\cite{sohoni2020no} and is even as competitive as group DRO~\cite{GroupDRO}.

\begin{figure}
	\begin{center}
	\includegraphics[width=0.9\linewidth]{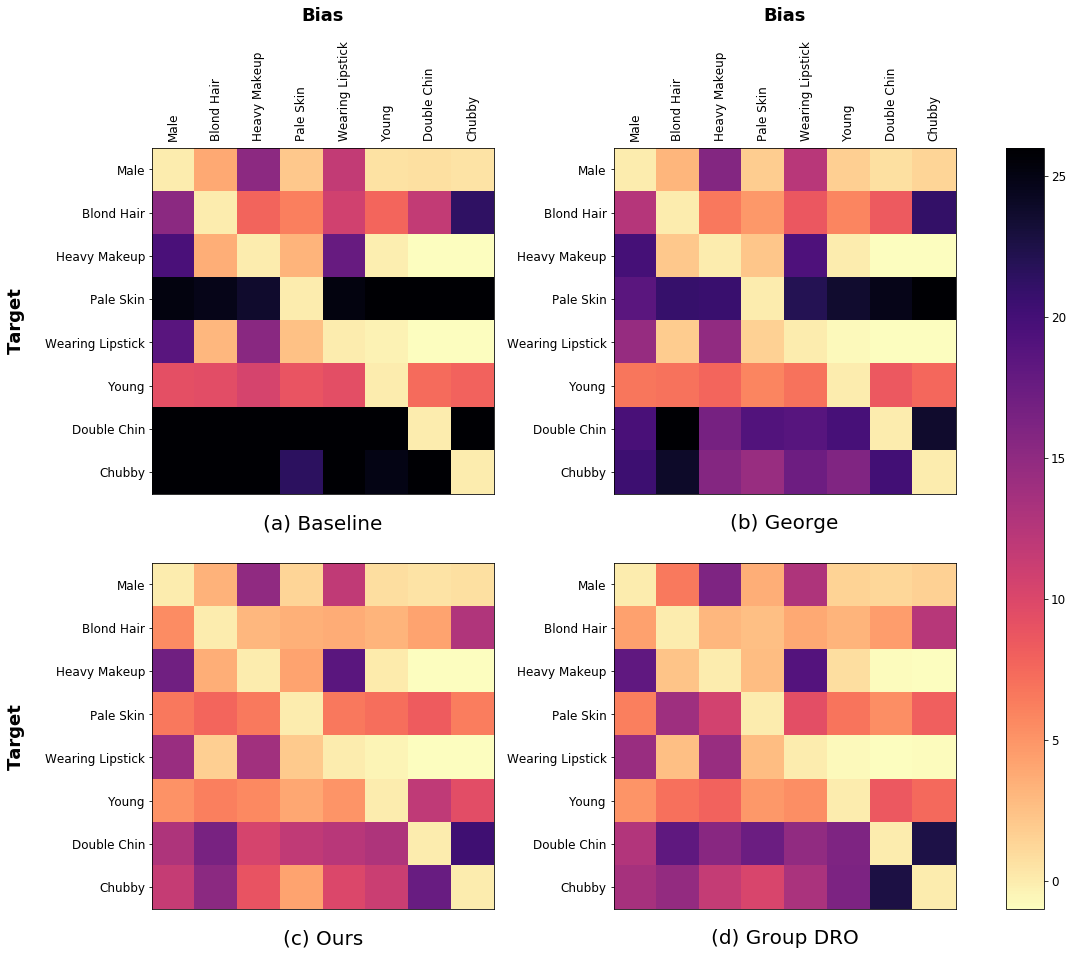}
	\end{center}
	\caption{
	Heatmap of the performance gap between overall accuracy and unbiased accuracy (\%p) with 4 different methods. 
	We use the same experimental setup with Figure~\ref{fig:heatmap}.
	The columns and rows denote bias and target attributes, respectively.
	In subfigure (a), the larger the performance gap, the more severe the algorithmic bias.
    As shown in the figure, even for the same target attribute, the gap varies largely depending on bias attributes. 
	Subfigure (b), (c) and (d) demonstrate that all methods mitigate the algorithmic bias,  while our approach is more effective than George.
	}
	\label{fig:heatmap_bias}
\end{figure}

\paragraph{Algorithmic bias with various bias attributes}
In Figure~\ref{fig:heatmap_bias}, we visualize the performance gap between overall accuracy and unbiased accuracy for each method to analyze the degree of algorithmic bias between target and bias attributes.
We use the same experimental setting with Figure~\ref{fig:heatmap}.
The larger the performance gap, the more severe the algorithmic bias.
This implies that, based on the performance gap from Figure~\ref{fig:heatmap_bias} (a), we can measure the existence of algorithmic bias on the CelebA dataset, \eg, the target attribute \textit{Heavy Makeup} is spuriously correlated to \textit{Male} and \textit{Wearing Lipstick} biases while not to \textit{Young}, \textit{Double Chin} and \textit{Chubby} biases.\footnote{As in the main paper, we suppose that the algorithmic
bias exists between target and bias attributes when a baseline model gives a large performance gap between its overall accuracy and unbiased accuracy (e.g., $>$ 5\% points).}
As shown in the figure, even with the same target attribute, the gap varies largely depending on bias attributes. 
We also observe that the algorithmic bias does not exist symmetrically, \eg, the target attribute \textit{Chubby} is spuriously correlated to \textit{Heavy Makeup} bias, not vice versa.
Compared to the baseline, all methods reduce the algorithmic bias while our framework is more effective than George~\cite{sohoni2020no} and as competitive as group DRO~\cite{GroupDRO}.

\paragraph{Multi-target classification}
We tested our framework with another setting, called multi-target classification, where a single backbone model adopts multiple classification heads.
To this end, we attached multiple linear classification layers, which correspond to individual targets, respectively, to a shared feature extractor.
For evaluation, we calculate unbiased accuracy for each target attribute separately, where the bias attribute is fixed to \textit{gender}.
Table~\ref{tab:multi_target} presents the multi-target classification results with several target attribute pairs, where our model achieves consistently better results than the compared unsupervised method in terms of unbiased accuracy, while it is as competitive as group DRO~\cite{GroupDRO}.

\begin{table*}[t]
\begin{center}
\caption{Unbiased accuracy~(\%) with multi-target classification scenario.
In this setting, each model is trained to classify multiple attributes jointly by adopting separate linear branches.
The bias attribute is fixed to \textit{Male}.
We report the unbiased accuracy for each target attribute separately.
}
\vspace{-0.cm}
\label{tab:multi_target}
\scalebox{0.85}{
\setlength\tabcolsep{8pt} \hspace{-0.2cm}
\hspace{-0.2cm}
\begin{tabular}{c|ccc|c}
\toprule
&\multicolumn{3}{|c|}{Unsupervised} & Supervised  \\
Targets & Baseline & George~\cite{sohoni2020no} & {{Ours}} & Group DRO~\cite{GroupDRO} \\
\hline
Blond Hair / Heavy Makeup & 78.92 / 71.46 & 83.06 / 70.99 & {\underline{89.78}} / {\bf72.25} & {\bf{90.38}} / {\underline{70.94}} \\
Blond Hair / Wearing Lipstick & 80.76 / 71.91 & 82.08 / 73.06 & {\bf89.09} / {\underline{77.34}} & {\underline{88.86}} / {\bf78.45}\\
Straight Hair / Oval Face & 69.93 / 60.84 & {\underline{76.77}} / 63.84 & {\bf78.33} / {\bf64.85} & 76.38 / {\underline{64.77}}\\
Straight Hair / Big Lips  & 70.05 / 60.14 & 69.73 / 63.22 & {\underline{76.03}} / {\bf66.39} & {\bf77.06} / {\underline{64.07}} \\
Blurry / Pale Skin & 73.89 / 68.18 & 79.51 / 79.45 & {\underline{87.35}} / {\bf89.28} & {\bf87.82} / {\underline{85.91}} \\
Blurry / Young  & 76.48 / 77.82 & 74.66 / 77.16 & {\bf88.64} / {\bf82.79} & {\underline{88.57}} / {\underline{82.14}} \\
\bottomrule
\end{tabular}
}
\end{center}
\vspace{-0.3cm}
\end{table*}

\end{document}